\documentclass[normalheadings]{scrartcl}
\KOMAoptions{
    paper=a4,
    fontsize=11pt,
    cleardoublepage=empty,
    footinclude=true
}

\usepackage{fullpage}
\usepackage{xspace}
\usepackage{color}
\usepackage{subfigure}
\usepackage{amsmath,amsthm,amssymb}
\usepackage{multirow}
\usepackage{tikz}
\usetikzlibrary{arrows,positioning,shapes.misc,shapes.geometric}
\usetikzlibrary{plotmarks}

\setkomafont{title}{\rmfamily}
\setkomafont{section}{\rmfamily\Large}
\setkomafont{subsection}{\rmfamily\large}
\setkomafont{subsubsection}{\rmfamily}

\newtheorem{proposition}{Proposition}
\newtheorem{conjecture}{Conjecture}

\theoremstyle{definition}
\newtheorem{definition}{Definition}
\newtheorem{example}{Example}

\tikzset{
    >=stealth',
    args/.style={circle,draw=black}
}

\renewcommand{\iff}{\ensuremath{\mathit{iff}}}						
\newcommand{\AF}{\ensuremath{\mathsf{AF}}\xspace}						
\newcommand{\arguments}{\ensuremath{\mathsf{Arg}}\xspace}				
\newcommand{\attacks}{\ensuremath{\rightarrow}\xspace}				
\newcommand{\AFcomplete}{\ensuremath{\AF=(\arguments,\attacks)}\xspace}		
\newcommand{\cA}{\ensuremath{\mathcal{A}}}
\newcommand{\cB}{\ensuremath{\mathcal{B}}}
\newcommand{\cC}{\ensuremath{\mathcal{C}}}
\newcommand{\argin}{\ensuremath{\mathsf{in}}}		
\newcommand{\argout}{\ensuremath{\mathsf{out}}}		
\newcommand{\argundec}{\ensuremath{\mathsf{undec}}}		

\newcommand{\cd}{\ensuremath{\,|\,}}
\newcommand{\ol}[1]{\ensuremath{\overline{#1}}}
\newcommand{\sat}{\ensuremath{\mathsf{sat}}}	
\newcommand{\atoms}{\ensuremath{\mathsf{At}}}	
\newcommand{\lang}[1]{\ensuremath{\mathcal{L}}_{#1}}	
\newcommand{\langc}[1]{\ensuremath{(\mathcal{L}_{#1}\cd\mathcal{L}_{#1})}}	
\newcommand{\image}[1]{\ensuremath{\mathsf{Im}\;#1}}

\newcommand{\rank}{\lambda}
\newcommand{\ranks}[1]{O_{#1}}
\newcommand{\Rank}{\Lambda}
\newcommand{\naturals}{\mathbb{N}}
\newcommand{\cO}{\mathcal{O}}
\newcommand{\Ostrat}[1]{\cO^{strat}_{#1}}
\newcommand{\WCom}{\mathit{WCom}}
\newcommand{\BF}{\ensuremath{\mathsf{BF}}\xspace}						

\newcommand{\Irstar}{\textbf{(Ir$^*$)}}
\newcommand{\VPstar}{\textbf{(VP$^*$)}}
\newcommand{\WVPstar}{\textbf{(WVP$^*$)}}
\newcommand{\DPstar}{\textbf{(DP$^*$)}}
\newcommand{\QPstar}{\textbf{(QP$^*$)}}

\newcommand{\cD}{\ensuremath{\mathcal{D}}}
\title{Stratified Labelings for Abstract Argumentation (Preliminary Report)}
\author{Matthias Thimm$^{\dagger,1}$, Gabriele Kern-Isberner$^{\ddagger,2}$\\~\\[-1ex] \small $^{\dagger}\texttt{thimm@uni-koblenz.de}$, $^{\ddagger}\texttt{gabriele.kern-isberner@cs.uni-dortmund.de}$\\ \small$^{1}$Institute for Web Science and Technologies, University of Koblenz, Germany\\[-1ex]\small$^{2}$Department of Computer Science, TU Dortmund, Germany}

\date{\today}

\begin{document}

\maketitle

\begin{abstract}
	We introduce stratified labelings as a novel semantical approach to abstract argumentation frameworks. Compared to standard labelings, stratified labelings provide a more fine-grained assessment of the \emph{controversiality} of arguments using ranks instead of the usual labels ``in'', ``out'', and ``undecided''. We relate the framework  of stratified labelings to conditional logic and, in particular, to the System Z ranking functions.
\end{abstract}

\section{Introduction}
Computational models of argumentation \cite{Rahwan:2009} are non-monotonic reasoning mechanisms that focus on the interplay of arguments and counterarguments. An argument is an entity that represents some grounds to believe in a certain statement and that can be in conflict with arguments establishing contradictory claims. The most commonly used framework to talk about general issues of argumentation is that of abstract argumentation \cite{Dung:1995}. In abstract argumentation, arguments are represented as atomic entities and the interrelationships between different arguments are modeled using an attack relation. Abstract argumentation has been thoroughly investigated in the past fifteen years and there is quite a lot of work on particularly semantical issues \cite{Baroni:2005,Caminada:2006,Baroni:2010,Wu:2010}. Several different kinds of semantics for abstract argumentation frameworks have been proposed that highlight different aspects of argumentation. Usually, semantics are given to abstract argumentation frameworks in terms of extensions or, more recently, labelings. For a specific labeling an argument is either accepted, not accepted, or undecided. In a fixed semantical context, there is usually a set of labelings that is consistent with the semantical context. In order to reason with a semantics one has to take either a credulous or skeptical perspective. That is, an argument is ultimately accepted wrt.\ a semantics if the argument is accepted by at least one labeling consistent with that semantics (the credulous perspective) or if the argument is accepted by all labelings consistent with the semantics (the skeptical perspective). 

In this paper we present a novel approach to assign semantics to abstract argumentation frameworks. We introduce \emph{stratified labelings} as a means to provide a graded assessment to arguments. A stratified labeling assigns to each argument of an argumentation framework some natural number (or infinity) which is meant to be interpreted as a degree of \emph{conflict}.
Our approach differs from other approaches to \emph{weighted} semantics such as probabilistic approaches \cite{Li:2011,Thimm:2012,Hunter:2013}, fuzzy approaches \cite{Janssen:2008}, or other weighted approaches \cite{Dunne:2011,Matt:2008,AmgoudBenNaimSUM2013} in this particular aspect. While those works usually interpret the weight/probability of an argument with the \emph{strength} of the argument, i.\,e., the larger the value the stronger the argument can be believed in, we interpret the ranking values as a measure of controversiality, i.\,e., the larger the value of an argument the more \emph{controversial} the argument can be seen. More specifically, if an argument is classified as ``in'' in the classical semantics, it usually gets a large value in those works and a low value in our work. If an argument is classified as ``out'' it usually gets a small value in other works and in our work as well (an argument that is \emph{clearly} ``out'' is not controversial). And an argument that is classified as ``undecided'' usually gets an intermediate value in other works while here it gets a large value, depending on the level of controversiality. 

This paper reports on preliminary work on the notion of stratified labelings and provides some first insights and comparisons with other works. In particular, we relate stratified labelings to the concept of ranking functions \cite{Spohn:1988,Goldszmidt:1996}. Ranking functions such as the Z-ordering of \cite{Goldszmidt:1996} (System Z) are used to provide semantics for conditional logics \cite{Nute:2002}. We show that stratified labelings for abstract argumentation and ranking functions for conditional logic are similar concepts, thus providing a conceptual bridge between the defeasible reasoning approaches of argumentation and conditional logic. In a preliminary fashion we also provide comparisons to further related works from argumentation theory.

The rest of this paper is organized as follows. In Section~\ref{sec:aa} we briefly review abstract argumentation frameworks and continue in Section~\ref{sec:strat} with presenting our semantical approach of stratified labelings. In Section~\ref{sec:rank} we relate stratified labelings to ranking-based reasoning approaches in conditional logics. In Section~\ref{sec:rel:amgoud} we compare our work with the ranking-based semantics for abstract argumentation presented in \cite{AmgoudBenNaimSUM2013}. In Section~\ref{sec:rel:char} we provide a comparison with the $(\sigma,\mathcal{U})$-characteristic of \cite{Baumann:2012}. In Section~\ref{sec:related} we provide comparisons to further related works and in Section~\ref{sec:summary} we conclude.

\section{Abstract Argumentation}\label{sec:aa}
\emph{Abstract argumentation frameworks} \cite{Dung:1995} take a very simple view on argumentation as they do not presuppose any internal structure of an argument. Abstract argumentation frameworks only consider the interactions of arguments by means of an attack relation between arguments.
\begin{definition}[Abstract Argumentation Framework]
	An \emph{abstract argumentation framework} $\AF$ is a tuple $\AF=(\arguments,\attacks)$ where \arguments is a set of arguments and \attacks is a relation $\attacks\subseteq \arguments\times\arguments$.
\end{definition}
For two arguments $\cA,\cB\in\arguments$ the relation $\cA \attacks \cB$ means that argument $\cA$ attacks argument $\cB$. Abstract argumentation frameworks can be concisely represented by directed graphs, where arguments are represented as nodes and edges model the attack relation.
\begin{example}\label{Example:AF1}
	Consider the abstract argumentation framework $\AF=(\arguments,\attacks)$ depicted in Figure~\ref{fig:AF1}. Here it is $\arguments=\{\cA_{1},\cA_{2},\cA_{3},\cA_{4},\cA_{5}\}$ and $\attacks=\{(\cA_{1},\cA_{2}),(\cA_{2},\cA_{1}),(\cA_{2},\cA_{3}),(\cA_{3},\cA_{4}),$ $(\cA_{4},\cA_{5}),(\cA_{5},\cA_{4}),$ $(\cA_{5},\cA_{3})\}$.	
	\begin{figure}[h]
		\begin{center}
			\begin{tikzpicture}[node distance=0.7cm]
			
				\node[args](args1){$\cA_{1}$};
				\node[args, right=of args1](args2){$\cA_{2}$};
				\node[args, right=of args2](args3){$\cA_{3}$};
				\node[args, right=of args3, yshift=0.7cm](args4){$\cA_{4}$};
				\node[args, right=of args3, yshift=-0.7cm](args5){$\cA_{5}$};
			
				\path(args1) edge [->,bend left] (args2);
				\path(args2) edge [->,bend left] (args1);
				\path(args2) edge [->] (args3);
				\path(args3) edge [->] (args4);
				\path(args4) edge [->, bend left] (args5);
				\path(args5) edge [->, bend left] (args4);
				\path(args5) edge [->] (args3);

			\end{tikzpicture}
		\end{center}
		\caption{A simple argumentation framework}
		\label{fig:AF1}
	\end{figure}
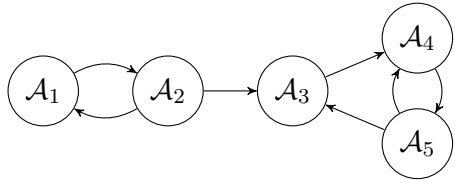
\end{example}
Semantics are usually given to abstract argumentation frameworks by means of extensions \cite{Dung:1995} or labelings \cite{Wu:2010}. An \emph{extension} $E$ of an argumentation framework $\AF=(\arguments,\attacks)$ is a set of arguments $E\subseteq\arguments$ that gives some coherent view on the argumentation underlying \AF. A labeling $L$ is a function $L:\arguments\rightarrow \{\argin,\argout,\argundec\}$ that assigns to each argument $\cA\in\arguments$ either the value \argin, meaning that the argument is accepted, \argout, meaning that the argument is not accepted, or \argundec, meaning that the status of the argument is undecided. Let $\argin(L)=\{\cA\mid L(\cA)=\argin\}$ and $\argout(L)$ resp.\ $\argundec(L)$ be defined analogously. As extensions can be characterized by the arguments that labeled $\argin$ in some labeling, we restrain our attention to labelings henceforth. In order to distinguish extension- and labeling-based semantics to the probabilistic semantics in the next section we denote the former \emph{classical semantics}.

In the literature \cite{Dung:1995,Caminada:2006} a wide variety of different types of classical semantics has been proposed. Arguably, the most important property of a semantics is its admissibility.
\begin{definition}
A labeling $L$ is called \emph{admissible} if and only if for all arguments $\cA\in\arguments$
\begin{enumerate}
	\item if $L(\cA)=\argout$ then there is $\cB\in\arguments$ with $L(\cB)=\argin$ and $\cB\attacks\cA$, and
	\item if $L(\cA)=\argin$ then $L(\cB)=\argout$ for all $\cB\in\arguments$ with $\cB\attacks \cA$,
\end{enumerate}
and it is called \emph{complete} if, additionally, it satisfies
\begin{enumerate}
	\setcounter{enumi}{2}
	\item if $L(\cA)=\argundec$ then there is no $\cB\in\arguments$ with $\cB\attacks\cA$ and $L(\cB)=\argin$ and there is a $\cB'\in\arguments$ with $\cB'\attacks\cA$ and $L(\cB')\neq\argout$.
\end{enumerate}
\end{definition}
The intuition behind admissibility is that an argument can only be accepted if there are no attackers that are accepted and if an argument is not accepted then there has to be some reasonable grounds. The idea behind the completeness property is that the status of argument is only \argundec\ if it cannot be classified as $\argin$ or $\argout$. Different types of classical semantics can be phrased by imposing further constraints.

\begin{definition}\label{def:semantics}
	Let \AFcomplete be an abstract argumentation framework and $L:\arguments\rightarrow \{\argin,\argout,\argundec\}$ a complete labeling.
	\begin{itemize}
		\item $L$ is \emph{grounded} if and only if $\argin(L)$ is minimal.
		\item $L$ is \emph{preferred} if and only if $\argin(L)$ is maximal.
		\item $L$ is \emph{stable} if and only if $\argundec(L)=\emptyset$.
		\item $L$ is \emph{semi-stable} if and only if $\argundec(L)$ is minimal.
	\end{itemize}
	All statements on minimality/maximality are meant to be with respect to set inclusion.
\end{definition}

Note that a grounded labeling is uniquely determined and always exists \cite{Dung:1995}. Besides the above mentioned types of classical semantics there are a lot of further proposals such as \emph{CF2 semantics} \cite{Baroni:2005}. However, in this paper we focus on complete (c), grounded (gr), preferred (p), stable (s), and semi-stable (ss) semantics. In the following, let $\sigma\in\{$c, gr, p, s, ss$\}$ be some semantics.

\begin{example}\label{Example:AF2}
	We continue Example~\ref{Example:AF1}. Consider the labeling $L$ defined via
	\begin{align*}
		L(\cA_{1}) &=\argin	&	L(\cA_{2}) &=\argout & L(\cA_{3}) &=\argout\\
		L(\cA_{4}) &=\argout &	L(\cA_{5}) &=\argin
	\end{align*}
	Clearly, $L$ is an admissible labeling as it satisfies properties 1.) and 2.) from above. Additionally, it is complete and also preferred, stable, and semi-stable. Furthermore, consider the labeling $L'$ defined via 
	\begin{align*}
		L'(\cA_{1}) &=\argout	&	L'(\cA_{2}) &=\argin & L'(\cA_{3}) &=\argout\\
		L'(\cA_{4}) &=\argin &	L'(\cA_{5}) &=\argout
	\end{align*}	
	The labeling $L'$ is also admissible, complete, preferred, stable, and semi-stable. Note, that the grounded labeling $L_{g}$ is defined via $L_{g}(\cA_{1})=L_{g}(\cA_{2})=L_{g}(\cA_{3}) =L_{g}(\cA_{4})=L_{g}(\cA_{5}) =\argundec$.
\end{example}

\section{Stratified Labelings}\label{sec:strat}
In the following, we define stratified labelings as a novel approach to give semantics to an abstract argumentation framework $\AFcomplete$. 

\begin{definition}\label{def:stratlabeling}
	Let $\AFcomplete$ be an abstract argumentation framework and let $\sigma$ be a semantics. A \emph{$\sigma$-stratified labeling} $S$ for $\AF$ is a function $S:\arguments\rightarrow\mathbb{N}\cup\{\infty\}$  such that there is a $\sigma$-labeling $L$ for $\AF$ and
	\begin{enumerate}
		\item if $\argin(L)=\emptyset$ then $S(\cA) =\infty$ for all $\cA\in \arguments$.
		\item if $\argin(L)\neq\emptyset$ then there is a $\sigma$-stratified labeling $S'$ for $\AF'=(\arguments',\attacks\cap\arguments'\times \arguments')$ with $\arguments'=\arguments\setminus \argin(L)$ such that 
		\begin{enumerate}		
			\item $S(\cA)=0$ for all $\cA\in\argin(L)$ and
			\item $S(\cA)=1+S'(\cA)$ for all $\cA\in \arguments\setminus\argin(L)$.
		\end{enumerate}
	\end{enumerate}
	A $\sigma$-stratified labeling $S$ is called \emph{finite} if $S^{-1}(\infty)=\emptyset$.
\end{definition}

The idea behind $\sigma$-stratified labelings is to measure the amount of \emph{controversiality} or \emph{indeterminateness} of assigning the label $\argin$ to an argument. In particular, a value $S(\cA)=0$ means that an argument is uncontroversially accepted. The larger the value the more controversial an argument becomes. Note that, in particular, there may be arguments which are considered ``out'' by the initial $\sigma$-labeling $L$ but classified with rank one by a corresponding stratified labeling while ``undecided'' arguments may get even larger values. This behavior is in contrast to other approaches for graded assessments of arguments \cite{Li:2011,Thimm:2012,Hunter:2013,Janssen:2008,Dunne:2011,Matt:2008,AmgoudBenNaimSUM2013} where controversial arguments are usually assessed as in between ``in'' and ``out'' arguments. The rationale behind the assessment of stratified labelings is that arguments classified as ``out'' with classical semantics are less controversial than undecided arguments (although they are not accepted they are uncontroversially classified as ``out''). The interpretation of stratified labelings follows the idea of dynamics of argumentation frameworks \cite{Cayrol:2008,Falappa:2009} and, specifically, the notion of \emph{enforcement} \cite{Baumann:2012}: how much must an argumentation framework be changed in order to accept a given argument? Arguments uncontroversially classified as ``out'' are (basically) more easily enforced. We will have another look at these issues in Section~\ref{sec:rel:char}.

Consider the following examples.

\begin{example}\label{Example:GSL1}
	The grounded-stratified labeling for the argumentation framework from Example~\ref{Example:AF1} is $S^{gr}_{\AF}$ with
	\begin{align*}
		S^{gr}_{\AF}(\cA_{1}) = S^{gr}_{\AF}(\cA_{2}) = S^{gr}_{\AF}(\cA_{3}) = S^{gr}_{\AF}(\cA_{4}) = S^{gr}_{\AF}(\cA_{5}) = \infty
	\end{align*}
\end{example}

\begin{example}\label{Example:GSL2}
	The grounded-stratified labeling for the argumentation framework depicted in Figure~\ref{fig:GSL2} is $S^{gr}_{\AF}$ with
	\begin{align*}
		S^{gr}_{\AF}(\cA_{1}) &= 0 &
		S^{gr}_{\AF}(\cA_{2}) &= 1 &
		S^{gr}_{\AF}(\cA_{3}) &= 2
	\end{align*}
	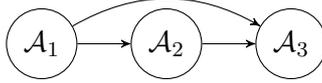
\begin{figure}[h]
		\begin{center}
			\begin{tikzpicture}[node distance=0.7cm]
			
				\node[args](args1){$\cA_{1}$};
				\node[args, right=of args1](args2){$\cA_{2}$};
				\node[args, right=of args2](args3){$\cA_{3}$};
			
				\path(args1) edge [->] (args2);
				\path(args1) edge [->,bend left] (args3);
				\path(args2) edge [->] (args3);
			\end{tikzpicture}
		\end{center}
		\caption{Argumentation framework from Example~\ref{Example:GSL2}}
		\label{fig:GSL2}
	\end{figure}
	The grounded labeling of $\AF$ assigns to $\cA_{1}$ the value $\argin$ and to all other arguments the value $\argout$. Therefore, $\cA_{1}$ gets the value $0$. Removing $\cA_{1}$ from $\AF$ yields a framework consisting of arguments $\cA_{2},\cA_{3}$ and $\cA_{2}$ attacking $\cA_{3}$. The grounded labeling of this framework assigns to $\cA_{2}$ the value $\argin$ and to $\cA_{3}$ the value $\argout$. Therefore, $\cA_{2}$ gets the value $1$. Finally, $\cA_{3}$ gets the value $2$.
\end{example}

\begin{example}\label{Example:GSL3}
	The grounded-stratified labeling for the argumentation framework depicted in Figure~\ref{fig:GSL3} is $S^{gr}_{\AF}$ with
	\begin{align*}
		S^{gr}_{\AF}(\cA_{1}) &= 0 &
		S^{gr}_{\AF}(\cA_{2}) &= 1 &
		S^{gr}_{\AF}(\cA_{3}) &= 3 \\
		S^{gr}_{\AF}(\cA_{4}) &= 1 &
		S^{gr}_{\AF}(\cA_{5}) &= 2
	\end{align*}
	\begin{figure}[h]
		\begin{center}
			\begin{tikzpicture}[node distance=0.7cm]
			
				\node[args](args1){$\cA_{1}$};
				\node[args, right=of args1](args2){$\cA_{2}$};
				\node[args, right=of args2](args3){$\cA_{3}$};
				\node[args, right=of args3, yshift=0.7cm](args4){$\cA_{4}$};
				\node[args, right=of args3, yshift=-0.7cm](args5){$\cA_{5}$};
			
				\path(args1) edge [->] (args2);
				\path(args2) edge [->] (args3);
				\path(args3) edge [->, bend left] (args4);
				\path(args4) edge [->, bend left] (args3);
				\path(args4) edge [->] (args5);
				\path(args5) edge [->] (args3);

			\end{tikzpicture}
		\end{center}
		\caption{Argumentation framework from Example~\ref{Example:GSL3}}
		\label{fig:GSL3}
	\end{figure}
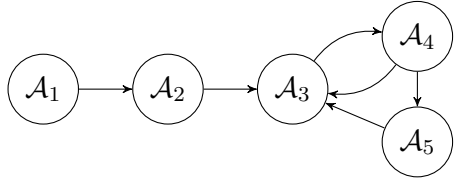
\end{example}
The last example also shows the advantage of using stratified labelings instead of ordinary labelings. While for $\AF$ from Example~\ref{Example:GSL3} only argument $\cA_{1}$ is labelled $\argin$ (with respect to grounded semantics), $\cA_{2}$ is labeled $\argout$, and all other arguments are labeled $\argundec$, the grounded-stratified labeling gives a more graded assessment of the arguments' controversiality.

\begin{example}\label{Example:SSL1}
	The argumentation framework $\AF$ shown in Figure~\ref{fig:SSL1} has six different stable-stratified labelings 
	\begin{align*}
		S^{s,1}_{\AF}(\cA_{1}) &= 0 &		S^{s,1}_{\AF}(\cA_{2}) &= 1 &		S^{s,1}_{\AF}(\cA_{3}) &= 2 \\
		S^{s,2}_{\AF}(\cA_{1}) &= 0 &		S^{s,2}_{\AF}(\cA_{2}) &= 2 &		S^{s,2}_{\AF}(\cA_{3}) &= 1 \\
		S^{s,3}_{\AF}(\cA_{1}) &= 1 &		S^{s,3}_{\AF}(\cA_{2}) &= 2	&	S^{s,3}_{\AF}(\cA_{3}) &= 3 \\
		S^{s,4}_{\AF}(\cA_{1}) &= 1 &		S^{s,4}_{\AF}(\cA_{2}) &= 3 &		S^{s,4}_{\AF}(\cA_{3}) &= 2 \\
		S^{s,5}_{\AF}(\cA_{1}) &= 2 &		S^{s,5}_{\AF}(\cA_{2}) &= 1 &		S^{s,5}_{\AF}(\cA_{3}) &= 3 \\
		S^{s,6}_{\AF}(\cA_{1}) &= 2 &		S^{s,6}_{\AF}(\cA_{2}) & = 3 &		S^{s,6}_{\AF}(\cA_{3}) & = 1 
	\end{align*}
	\begin{figure}[h]
		\begin{center}
			\begin{tikzpicture}[node distance=0.7cm]
			
				\node[args](args1){$\cA_{1}$};
				\node[args, right=of args1](args2){$\cA_{2}$};
				\node[args, below=of args2](args3){$\cA_{3}$};

				\path(args1) edge [->] (args2);
				\path(args1) edge [->] (args3);
				\path(args2) edge [->] (args1);
				\path(args2) edge [->] (args3);
				\path(args3) edge [->] (args1);
				\path(args3) edge [->] (args2);

			\end{tikzpicture}
		\end{center}	
		\caption{Argumentation framework from Example~\ref{Example:SSL1}}
		\label{fig:SSL1}
	\end{figure}
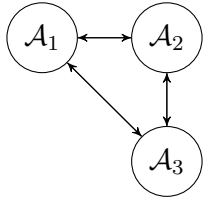	
\end{example}

We now look at some general properties of stratified labelings.

\begin{proposition}
	Let $\AFcomplete$ be an argumentation framework.
	\begin{enumerate}
		\item The grounded-stratified labeling $S^{gr}_{\AF}$ always exists and is uniquely determined.
		\item If $\cA\attacks\cA$ for some $\cA\in\arguments$ then $S(\cA)=\infty$ for every semantics $\sigma$ and $\sigma$-stratified labeling $S$.
		\item for every stable-stratified labeling $S$ it holds $\infty\notin \image{S}$.
	\end{enumerate}
	\begin{proof}
		Let $\AFcomplete$ be an argumentation framework.
		\begin{enumerate}
			\item Due to \cite{Dung:1995} there is exactly one grounded labeling $L^{gr}$ of $\AF$. It follows that for every two grounded-stratified labelings $S_{1},S_{2}$ it holds $S_{1}^{-1}(0)=S_{2}^{-1}(0)$. By induction, it follows $S_{1}^{-1}(i)=S_{2}^{-1}(i)$ for every $i\in\mathbb{N}\cup\{\infty\}$ and therefore $S_{1}=S_{2}$.
			\item If $\cA\attacks\cA$ then $L(\cA)\neq \argin$ for every semantics $\sigma$ and $\sigma$-labeling $L$. Therefore, neither condition 1.)\ nor condition 2.)\ in Definition~\ref{def:stratlabeling} can be satisfied for $\cA$ and it follows $S(\cA)=\infty$ for every semantics $\sigma$ and $\sigma$-stratified labeling $S$.
			\item Assume that $S$ is a stable-stratified labeling with $S(\cA)=\infty$ and let $L$ be the labeling in condition 3.)\ of Definition~\ref{def:stratlabeling} responsible for setting $S(\cA)=\infty$. As $L$ is stable (requiring $\argundec(L)=\emptyset$) and $L(\cA)\neq \argin$ it follows $L(\cA)=\argout$. Then there must be a $\cB$ with $\cB\attacks \cA$ and $L(\cB)= \argin$. This is a contradiction and therefore $\infty\notin \image{S}$.\qedhere
		\end{enumerate}		
	\end{proof}
\end{proposition}

\begin{proposition}
\label{prop_char_strat_labelings}
Each $\sigma$-stratified labeling $S$ of an argumentation framework $\AF=(\arguments,\attacks)$ is characterized by a set of nested subsets $A_0 \supseteq A_1 \supseteq \ldots \supseteq A_k \supseteq A_{-1}$ of $\arguments$ with $k \geq -1$, $A_0 =\arguments$ if $k \geq 0$, 
and an appertaining vector $(L_0, L_1, \ldots, L_k, L_{-1})$ of $\sigma$-labelings $L_i$ such that 
\begin{enumerate}
\item $L_i$ is a labeling on $(A_i, \attacks \cap (A_i \times A_i))$, $-1 \leq i \leq k$, 
\item $\argin(L_i) = A_i \backslash A_{i+1} \neq \emptyset$, $0 \leq i \leq k$, 
\item  $A_{-1} = \arguments \backslash (\cup_{i=0}^k A_i)$, $\argin(L_{-1}) = \emptyset$, 
\item $S(\cA) = max \{i \mid \cA \in A_i,  0 \leq i \leq k \}$, if $k \geq 0$, 
\item $S(\cA) = S(\cB)$ iff $\cA, \cB \in A_{i}$ for some $-1 \leq i \leq k$, 
\item $S(\cA) \leq S(\cB)$ iff $\cA\in A_{i}$ implies $\cB\in A_{i}$ for all $-1 \leq i \leq k$, 
\item $S(\cA) = \infty$ for all $\cA \in A_{-1}$.
\end{enumerate}
If $S$ is characterized  as given  above, we write $S \leftrightarrow \langle (A_0, A_1, \ldots, A_k, A_{-1}), (L_0, L_1, \ldots, L_k, L_{-1})\rangle$. 
\end{proposition}
Note that $k=-1$ is possible, in which case we have $A_{-1} = \arguments$, and that on the other hand, $A_{-1}$ can be empty, which is equivalent to $S$ being finite. 

\section{Relating Stratified Labelings with Ranking Functions}\label{sec:rank}
In the following, we relate stratified labelings with ranking functions from conditional reasoning \cite{Spohn:1988,Goldszmidt:1996}. For that, we first give some background information on conditionals and ranking functions in Section~\ref{subsec:prelim:cond} and provide our comparative analysis in Section~\ref{subsec:comp:rank}.
\subsection{Conditionals and Ranking Functions}\label{subsec:prelim:cond}
Let $\atoms$ be a set of propositional atoms and $\lang{\atoms}$ the propositional language generated using the usual connectives. Let $\Omega_{\atoms}$ be the set of interpretations of $\lang{\atoms}$ and $\models$ the standard propositional satisfaction relation.

\begin{definition}
	A \emph{conditional} $\delta$ has the form $\delta=(\phi\cd\psi)$ with $\phi,\psi\in\lang{\atoms}$. Let $\langc{\atoms}$ denote the set of all conditionals.
\end{definition}
A $(\phi\cd\psi)$ is a defeasible rule which states that $\psi$ usually/defeasibly implies $\phi$.
If $\psi\equiv\top$ we write $(\phi)$ instead of $(\phi\cd\psi)$. An interpretation $\omega\in\Omega_{\atoms}$
\begin{itemize}
	\item \emph{verifies} a conditional $(\phi\cd\psi)$ if $\omega\models \phi\psi$,
	\item \emph{falsifies} a conditional $(\phi\cd\psi)$ if $\omega\models \ol{\phi}\psi$,
	\item \emph{satisfies} a conditional $(\phi\cd\psi)$ if it does not falsify it.
\end{itemize}

\begin{definition}
	A \emph{knowledge base} $\Delta$ is a finite set of conditionals $\Delta\subseteq\langc{\atoms}$.
\end{definition}
An interpretation $\omega\in\Omega_{\atoms}$ satisfies $\Delta$ if it satisfies every conditional in it. Define
\begin{align*}
	\sat_{\Delta}(\omega)& = \{\delta\in\Delta\mid \omega\text{~satisfies~}\delta\}
\end{align*}

Semantics can be given to (conditional) knowledge bases by means of ranking functions.

\begin{definition}
A \emph{ranking function} $\kappa$ is a function $\kappa: \Omega_{\atoms}\rightarrow \mathbb{N}\cup\{\infty\}$ with $\kappa^{-1}(0) \neq \emptyset$.
\end{definition}

A ranking function partitions the set of possible worlds into ordered layers. The intuition of the rank $\kappa(\omega)$ is that the larger the value the more implausible the interpretation is to be assessed. Interpretations at rank zero are considered to be most plausible. For $\phi\in\lang{\atoms}$ we write
\begin{align*}
	\kappa(\phi)=\left\{\begin{array}{ll}
					\min\{\kappa(\omega)\mid\omega\models\phi\} & \text{if~}\phi\not\models\perp\\
					\infty & \text{otherwise}
					\end{array}\right.
\end{align*}
A conditional $(\phi\cd\psi)\in\langc{\atoms}$ is \emph{accepted} by $\kappa$, denoted by $\kappa\models (\phi\cd\psi)$, if $\kappa(\phi\psi)<\kappa(\ol{\phi}\psi)$. This means, that from the perspective of $\kappa$ interpretations satisfying $\phi$ and $\psi$ are more plausible than interpretations satisfying $\psi$ but not $\phi$.
A knowledge base is consistent if there is a $\kappa$ that accepts all conditionals in $\Delta$.

For a specific knowledge base $\Delta$ there is usually an infinite number of ranking functions accepting all its conditionals. In order to allow for commonsense reasoning one usually focuses on a specific class or a single specific ranking function. One standard approach is the Z-ordering \cite{Goldszmidt:1996} which is based on the notion of \emph{tolerance}.

\begin{definition}
A conditional $(\psi\cd\phi)$ is \emph{tolerated} by $\Delta$ if there is a $\omega\in\Omega_{\atoms}$ such that $\omega$ verifies $(\psi\cd\phi)$ and satisfies $\Delta$.
\end{definition}

With the definition of tolerance one can partition the conditionals in $\Delta$ with respect to their \emph{compatibility} to the other conditionals.

\begin{definition}
	Let $\Delta$ be consistent. The $Z$-partitioning $(\Delta_{0},\ldots,\Delta_{n})$ of $\Delta$ is defined as
	\begin{enumerate}
		\item $\Delta_{0} = \{\delta\in\Delta\cd\Delta\text{~tolerates~}\delta\}$,
		\item $\Delta_{1},\ldots,\Delta_{n}$ is the $Z$-partitioning of $\Delta\setminus\Delta_{0}$.
	\end{enumerate}
\end{definition}
For $\delta\in\Delta$ define furthermore
\begin{align*}
	Z_{\Delta}(\delta) = i\qquad \iff \qquad\delta\in \Delta_{i}\text{~and~}(\Delta_{0},\ldots,\Delta_{n})\text{~is the $Z$-partitioning of~}\Delta
\end{align*}

Finally, the ranking function $\kappa_{\Delta}^{z}$ is defined as follows.
\begin{definition}
	Let $\Delta$ be consistent. The ranking function $\kappa_{\Delta}^{z}$ is defined via
	\begin{align*}
		\kappa_{\Delta}^{z}(\omega) & = \left\{\begin{array}{ll}
								0	& \text{if~}  \omega\text{~satisfies~}\Delta\\
								\max\{Z(\delta)\mid \omega \text{~falsifies~}\delta\}+1 & \text{otherwise}
							\end{array}\right.
	\end{align*}
\end{definition}

Reasoning with the ranking function $\kappa_{\Delta}^{z}$ satisfies many commonsense reasoning properties, see e.\,g.\ \cite{Goldszmidt:1996}.

\subsection{Stratified Labelings and Ranking Functions}\label{subsec:comp:rank}
We now turn to analyzing the similarities between stratified labelings and ranking functions. For that we show how any (conditional) knowledge base can be transformed into an abstract argumentation framework such that argumentative reasoning based on stratified labelings in this framework is equivalent to reasoning based on ranking functions on $\Delta$ itself.

Let $\Delta\subseteq \langc{\atoms}$ be a consistent knowledge base.

\begin{definition}
	Define the preference relation $\prec^{Z}$ on $\Omega_{\atoms}$ via $\omega_{1}\prec^{Z}\omega_{2}$ iff 
	\begin{quote}
		($\omega_{1}\models\Delta$ and $\omega_{2}\not\models\Delta$) or $\max\{Z_{\Delta}(\delta)\mid \omega_{1} \text{~falsifies~}\delta\}<\max\{Z_{\Delta}(\delta)\mid \omega_{2} \text{~falsifies~}\delta\}$
	\end{quote}
\end{definition}

Define the \emph{$\prec^{\Delta}_{Z}$-induced argumentation framework} $\AFcomplete$ of $\Delta$ via 
\begin{align*}
	\arguments & = \Omega_{\atoms}\\
	\attacks & = \{(\omega_{1},\omega_{2})\mid \omega_{1}\prec^{\Delta}_{Z}\omega_{2}\}
\end{align*}

\begin{proposition}
	Let $\Delta$ be a consistent knowledge base and let $\AF$ be its $\prec^{\Delta}_{Z}$-induced argumentation framework. Then $\kappa^{z}_{\Delta} = S^{gr}_{\AF}$.	
\end{proposition}

\begin{example}\label{ex:afocf1}
	Let $\Delta$ be a knowledge base given via
	\begin{align*}
		\Delta & = \{(b\cd p), (\ol{f}\cd p), (f\cd b)\}\quad.
	\end{align*}
	The ranking function $\kappa^{z}_{\Delta}$ is defined in Table~\ref{tb:afocf1}, cf.\ Table~\ref{tb:afocf1b} for an overview on which interpretation satisfies and verifies which conditional. The $\prec^{\Delta}_{Z}$-induced argumentation framework $\AF^{Z}_{\Delta}$ is depicted in Figure~\ref{fig:afocf1}. 
	\begin{table}[h]
		\begin{center}
		\begin{tabular}{c|c}
			$\Omega_{\atoms}=\arguments$	&	$\kappa^{z}_{\Delta} = S^{gr}_{\AF^{Z}_{\Delta}}$	\\
			\hline
			$pbf$			&	2\\
			$pb\ol{f}$			&	1\\
			$p\ol{b}f$			&	2\\
			$p\ol{b}\ol{f}$		&	2\\
			$\ol{p}bf$			&	0\\
			$\ol{p}b\ol{f}$		&	1\\
			$\ol{p}\ol{b}f$		&	0\\
			$\ol{p}\ol{b}\ol{f}$	&	0
		\end{tabular}
		\end{center}
		\caption{Ranking functions/grounded-stratified labelings of Example~\ref{ex:afocf1}}
		\label{tb:afocf1}
	\end{table}
	\begin{table}[h]
		\begin{center}
		\begin{tabular}{c|c|c|c|c|c|c}
			$\Delta$		&	\multicolumn{2}{|c|}{$(b\cd p)$} &	\multicolumn{2}{|c|}{$(\ol{f}\cd p)$}		& \multicolumn{2}{|c}{$(f\cd b)$}\\
						&	satisf.  & verif.				&	satisf.  & verif.						&	satisf.  & verif.\\
			\hline	
			$pbf$			&	X		&	X			&			&					&	X		&	X\\			
			$pb\ol{f}$			&	X		&	X			&	X		&	X				&			&	\\
			$p\ol{b}f$			&			&				&			&					&	X		&	\\
			$p\ol{b}\ol{f}$		&			&				&	X		&	X				&	X		&	\\
			$\ol{p}bf$			&	X		&				&	X		&					&	X		&	X\\
			$\ol{p}b\ol{f}$		&	X		&				&	X		&					&			&	\\
			$\ol{p}\ol{b}f$		&	X		&				&	X		&					&	X		&	\\
			$\ol{p}\ol{b}\ol{f}$	&	X		&				&	X		&					&	X		&	
		\end{tabular}
		\end{center}
		\caption{Conditional verification/satisfaction in Example~\ref{ex:afocf1}}
		\label{tb:afocf1b}
	\end{table}
	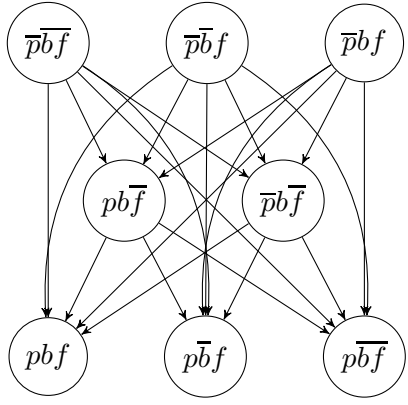
\begin{figure}[h]
		\begin{center}
			\begin{tikzpicture}[node distance=1cm]
				\node[args](pbf1){$\ol{p}\ol{b}\ol{f}$};
				\node[args, right=of pbf1](pbf2){$\ol{p}\ol{b}f$};
				\node[args, right=of pbf2](pbf3){$\ol{p}bf$};
				\node[args, below=of pbf1, xshift=1cm](pbf4){$pb\ol{f}$};
				\node[args, right=of pbf4](pbf5){$\ol{p}b\ol{f}$};
				\node[args, below=of pbf4, xshift=-1cm](pbf6){$pbf$};				
				\node[args, right=of pbf6](pbf7){$p\ol{b}f$};
				\node[args, right=of pbf7](pbf8){$p\ol{b}\ol{f}$};
				
				\path(pbf1) edge [->] (pbf4);
				\path(pbf1) edge [->] (pbf5);
				\path(pbf1) edge [->] (pbf6);
				\path(pbf1) edge [->, bend left] (pbf7);
				\path(pbf1) edge [->] (pbf8);

				\path(pbf2) edge [->] (pbf4);
				\path(pbf2) edge [->] (pbf5);
				\path(pbf2) edge [->, bend right] (pbf6);
				\path(pbf2) edge [->] (pbf7);
				\path(pbf2) edge [->, bend left] (pbf8);

				\path(pbf3) edge [->] (pbf4);
				\path(pbf3) edge [->] (pbf5);
				\path(pbf3) edge [->] (pbf6);
				\path(pbf3) edge [->, bend right] (pbf7);
				\path(pbf3) edge [->] (pbf8);
				
				\path(pbf4) edge [->] (pbf6);
				\path(pbf4) edge [->] (pbf7);
				\path(pbf4) edge [->] (pbf8);
				
				\path(pbf5) edge [->] (pbf6);
				\path(pbf5) edge [->] (pbf7);
				\path(pbf5) edge [->] (pbf8);
				
			\end{tikzpicture}
		\end{center}
		\caption{$\prec^{\Delta}_{Z}$-induced argumentation framework $\AF^{Z}_{\Delta}$ from Example~\ref{Example:GSL3}}
		\label{fig:afocf1}
	\end{figure}
\end{example}	

\section{Relating Stratified Labelings with Ranking-based semantics}\label{sec:rel:amgoud}
In \cite{AmgoudBenNaimSUM2013}, the authors consider ranking-based semantics of argumentation frameworks, i.\,e., they interpret  $\AF=(\arguments,\attacks)$ uniquely in terms of a total preorder on $\arguments$ that expresses acceptability. They set up a set of postulates that such semantics should satisfy, and present examples for ranking-based semantics as a proof of concept. 

In our approach, stratified labelings are not uniquely determined by the argumentation framework, but different labelings are possible, depending on the chosen semantics. To make the ideas of \cite{AmgoudBenNaimSUM2013} applicable, we first generalize their properties to handle classes of rankings for an argumentation framework. Furthermore, to comply better with the intuitive notion of rankings as a kind of (numerical) ordinal degrees, we specify rankings as an assignment of natural numbers to arguments. 

\begin{definition}
\label{def_ranking}
An \emph{ordinal ranking} $\rank$ of an argumentation framework $\AF=(\arguments,\attacks)$ is a function 
$\rank: \arguments \to \naturals \cup {\infty}$. If for $\cA, \cB \in \arguments$, $\rank(\cA) \leq \rank(\cB)$, then we say that $\cA$ is at least as acceptable as $\cB$.  
Let $\Rank(\AF)$ be the set of all ordinal rankings of $\AF$. 
\end{definition}
For the above rankings we consider the interpretation of values as given by \cite{AmgoudBenNaimSUM2013}, i.\,e., rankings express \emph{degrees of acceptability}: The lower the rank of an argument, the more acceptable it is deemed. If $\rank(\cA) = \infty$, then $\cA$ is not acceptable at all. Note that $\sigma$-stratified labelings are syntactically ordinal rankings but differ in their interpretation. More specifically, stratified labelings are meant as a measure of controversiality and not acceptability. Nonetheless, in the following we adopt the properties discussed in \cite{AmgoudBenNaimSUM2013} and apply them to stratified labelings as well. For that, it is clear that each ranking defined by Definition \ref{def_ranking} induces a ranking in the sense of \cite{AmgoudBenNaimSUM2013} and vice versa. Moreover, it would be possible to define stratified labelings as total preorders as well, but using natural numbers (plus $\infty$) allows a more compact handling of technical details. 

Here, rankings are not looked upon as transformations of argumentation frameworks as in \cite{AmgoudBenNaimSUM2013}  but are rather meant to be kind of models of argumentation frameworks. With the next definition, we define  semantics to argumentation frameworks by assigning to each framework a class of such ordinal rankings.

\begin{definition}
\label{def_ordinal_semantics}
An \emph{ordinal semantics} $\cO$ is a function that assigns a class of rankings to each argumentation framework $\AF=(\arguments,\attacks)$:
\[
\cO: \AF \mapsto \ranks{\AF} \subseteq \Rank(\AF).
\]
\end{definition}

$\sigma$-stratified labelings define an ordinal semantics for argumentation frameworks: 

\begin{definition}
\label{def_strat_ord_sem}
Let $\sigma$ be a semantics. The \emph{ordinal $\sigma$-stratified semantics} $\Ostrat{\sigma}$ is defined by 
\[
\Ostrat{\sigma}(\AF) \ = \ \{S \mid S \text{~is a~$\sigma$-stratified labelings for~} \AF\}.
\]
\end{definition}

Now, we elaborate on formal properties of ordinal semantics in analogy to \cite{AmgoudBenNaimSUM2013}, enhancing the names of the postulates with an asterisk to indicate that they refer to semantics in terms of sets of rankings.  

First, if two argumentation frameworks $\AF_1, \AF_2$ are isomorphic, then they should have basically the same semantics. We define isomorphisms between argumentation frameworks in terms of graph isomorphisms. 

\begin{definition}
\label{def_arg_iso}
Let $\AF_1 = (\arguments_1,\attacks_1), \AF_2 =(\arguments_2,\attacks_2)$ be two argumentation frameworks. 
An \emph{isomorphism} $\varphi$ from $\AF_1$ to $\AF_2$ is a bijective mapping $\phi: \arguments_1 \to \arguments_2$ such that for all $\cA, \cB \in  \arguments_1$, $\cA \attacks_1 \cB$ iff $\varphi(\cA) \attacks_1 \varphi(\cB)$. The frameworks  $\AF_1$ and $\AF_2$ are called \emph{isomorphic} if there is an isomorphism $\varphi$ from $\AF_1$ to $\AF_2$.
\end{definition}

The first property, \textbf{Abstraction$^*$} states that isomorphisms between two argumentation frameworks are apt to carry over ordinal semantics: 

\begin{description}
\item{\textbf{Abstraction$^*$ (Ab$^*$)}} An ordinal semantics $\cO$ satisfies \textbf{(Ab$^*$)} iff for any isomorphic argumentation frameworks $\AF_1 = (\arguments_1,\attacks_1)$ and $\AF_2 =(\arguments_2,\attacks_2)$, and for every isomorphism $\varphi: \AF_1 \to \AF_2$, it holds that 
\begin{equation}
\label{eq_abstraction}
\cO(\AF_2) = \{\rank_1 \circ \varphi^{-1} \mid \rank_1 \in \cO(\AF_1)\} = \cO(\AF_1) \circ \varphi^{-1}. 
\end{equation}
\end{description}

\begin{proposition}
\label{prop_strat_satisfies_ab}
$\Ostrat{\sigma}$ satisfies \textbf{(Ab$^*$)} for all semantics $\sigma$. 
\begin{proof}
Let $\varphi: \AF_1 \to \AF_2$ be an isomorphism. For each $\sigma$-stratified labeling $S_1$ on $\AF_1$ that is characterized by $A_0 \supseteq A_1 \supseteq \ldots \supseteq A_k \supseteq A_{-1}$ and $(L_0, L_1, \ldots, L_k, L_{-1})$, $(L_0 \circ \varphi^{-1}, L_1 \circ \varphi^{-1}, \ldots, L_k \circ \varphi^{-1}, L_{-1} \circ \varphi^{-1})$ defines a $\sigma$-stratified labeling $S_2$ on $\varphi(A_0) \supseteq \varphi(A_1) \supseteq \ldots \supseteq \varphi(A_k) \supseteq \varphi(A_{-1})$ such that $S_2 = S_1 \circ \varphi^{-1}$. 
\end{proof}
\end{proposition}

Also the next property, 
\textbf{Irrelevance$^*$ \Irstar}
deals merely with properties of the argumentation graph. Note that \textbf{Irrelevance$^*$ \Irstar} corresponds to \textbf{Independence} in \cite{AmgoudBenNaimSUM2013}. 

Let $\WCom(\AF)$ be the set of all subgraphs of $\AF$ that arise from (finite) unions of weakly connected components of $\AF$; in particular, each weakly connected component of $\AF$ is contained in $\WCom(\AF)$. Note that each $\BF \in \WCom(\AF)$ contains all relevant information for labelings, as it contains all relevant edges. We consider labelings and rankings on elements $\BF$ of $\WCom(\AF)$.

\begin{description}
\item{\textbf{Irrelevance$^*$ \Irstar}} An ordinal semantics $\cO$ satisfies \Irstar\ iff for all argumentation frameworks $\AF$ such that $\cO(\AF) \neq \emptyset$, and for any $\BF \in \WCom(\AF)$, for all $\lambda' \in \cO(\BF)$, there is $\lambda \in \cO(AF)$ such that, for any $\cB_1, \cB_2 \in \BF$, the following conditions are fulfilled: 
\begin{itemize}
\item[(i)] $\lambda'(\cB_1) = \lambda'(\cB_2)$ iff $\lambda(\cB_1) = \lambda(\cB_2)$, and 
\item[(ii)] $\lambda'(\cB_1) \leq \lambda'(\cB_2)$ iff $\lambda(\cB_1) \leq \lambda(\cB_2)$.
\end{itemize}
\end{description}

\begin{proposition}
\label{prop_strat_satisfies_ir}
$\Ostrat{\sigma}$ satisfies \textbf{(Ir$^*$)} for all semantics $\sigma$.
\begin{proof}
Let $\AF$ be an argumentation framework such that $\cO(\AF) \neq \emptyset$, let $\BF \in \WCom(\AF)$; then $\BF^c = \AF \backslash \BF$ is in $\WCom(\AF)$ as well. 
First, if $L$ is a $\sigma$-labeling on $\AF$, it can be partitioned into $\sigma$-labelings $(L', L'')$ with $\sigma$-labeling $L'$ on $\BF$ and $L''$ on $\BF^c$ such that 
\[
L(\cA) = \left\{
\begin{array}{l@{\ \mbox{if} \ }l}
L'(\cA) & \cA \in \BF,\\
L''(\cA) & \cA \in \BF^c,
\end{array}
\right.
\]
in particular, $\argin(L) = \argin(L') \cup \argin(L'')$. 
In principle, the same can be done for $\sigma$-stratified labelings $S$. If $S$ is characterized by $(A_0, A_1, \ldots, A_k, A_{-1})$ and appertaining $\sigma$-labelings $(L_0, L_1, \ldots, L_k,$ $L_{-1})$, then intersecting each $A_i$ with $\BF$ resp.\ $\BF^c$ gives rise to  labelings within the scope of $\BF$ resp.\  $\BF^c$. However, as it can be the case that $\argin(L_i) \cap \BF = \emptyset$ or $\argin(L_i) \cap \BF^c = \emptyset$, the final stratum $*_{-1}$ can be reached earlier. Due to the properties of the semantics, either both $\argin(L_i) \cap \BF $ or $\argin(L_i) \cap \BF^c$ are not empty for all $i$ (if $\sigma$ = stable), or if one of them is empty before the final stratum of $S$ is reached, all following intersections must also be empty while the $\argin(L_i)$ then are concentrated on the other component. The other way round,  $\sigma$-labelings on the components $\BF$ and $\BF^c$ can be combined to a $\sigma$-labeling on $\AF$. If these relationships hold between a $\sigma$-labeling $S$ on $\AF$ and $\sigma$-labeling $S'$ resp.\ $S''$ on $\BF$ resp.\ $\BF^c$, then we write $S = (S', S'')$, and $S' = S|_{\BF}$. 

Let $\sigma$ be a semantics and consider $\Ostrat{\sigma}$. Let $\AF$ be an argumentation framework such that $\cO(\AF) \neq \emptyset$, let $\BF \in \WCom(\AF)$, and let $S' \in \Ostrat{\sigma}(\BF)$ be a $\sigma$-stratified labeling of $\BF$, $S' \leftrightarrow \langle (B_0, B_1, \ldots, B_j, B_{-1}), (L'_0, L'_1, \ldots, L'_j, L'_{-1})\rangle$. From the construction above, and since $\cO(\AF) \neq \emptyset$, there is a $\sigma$-stratified labeling $S \in \Ostrat{\sigma}(\AF)$, $S \leftrightarrow \langle (A_0, A_1, \ldots, A_k, A_{-1}), (L_0, L_1, \ldots, L_k,$ $L_{-1})\rangle$   such that $S' = S|_{\BF}$, i.\,e., in particular, $j \leq k$, $B_i = A_i \cap \BF$, $L'_{i} = L_i|_{\BF}$.  For any two arguments $\cB_1, \cB_2 \in \BF$, due to Proposition \ref{prop_char_strat_labelings}, we have $S(\cB_1) = S(\cB_2)$ iff $\cB_1, \cB_2$ are elements of the same $A_i$, hence iff they are elements of the same $B_i$, therefore iff $S'(\cB_1) = S'(\cB_2)$. Similarly, $S(\cB_1) \leq S(\cB_2)$ iff $S'(\cB_1) \leq S'(\cB_2)$; note that $S(\cB) = \infty$ implies $S'(\cB) = \infty$, and $S(\cB), S'(\cB) \leq \infty$ for all arguments $\cB$. This completes the proof. 
\end{proof}
\end{proposition}

\begin{description}
\item{\textbf{Void Precedence$^*$ \VPstar}} An ordinal semantics $\cO$ satisfies \VPstar\ iff for all argumentation frameworks $\AF= (\arguments,\attacks)$, for all $\lambda \in \cO(\AF)$, for all $\cA, \cB \in \arguments$ the following holds: If $\cA$ is not attacked but $\cB$ is attacked, then $\lambda(\cA) < \lambda(\cB)$. 
\end{description}

\begin{proposition}
\label{prop_strat_not_satisfies_vp}
For every semantics $\sigma$, $\Ostrat{\sigma}$ does not satisfy \textbf{(VP$^*$)}.
\end{proposition}

This can be easily seen since for any $\sigma$, $\sigma$-labelings do not distinguish between the value ``in'' and not being attacked at all. And indeed, \VPstar\ is not indebatable because one might deem an argument that has survived attacks not to be worse than arguments that have not proved their strength against counterarguments. So, we propose a weakened version of \VPstar: 

\begin{description}
\item{\textbf{Weak Void Precedence$^*$ \WVPstar}} An ordinal semantics $\cO$ satisfies \WVPstar\ iff for all argumentation frameworks $\AF= (\arguments,\attacks)$, for all $\lambda \in \cO(\AF)$, for all $\cA, \cB \in \arguments$ the following holds: If $\cA$ is not attacked at all, then $\lambda(\cA) \leq \lambda(\cB)$. 
\end{description}

\WVPstar\ ensures that non-attacked arguments are at least as acceptable as any other arguments. 

\begin{proposition}
\label{prop_strat_satisfies_wvp}
$\Ostrat{\sigma}$ satisfies \textbf{(WVP$^*$)} for all semantics $\sigma$.
\begin{proof}
Let $\AF= (\arguments,\attacks)$ be an argumentation framework and let $\sigma$ be a semantics, let $\cA \in \arguments$ be an argument. If $\cA$ is not attacked at all, then for all $\sigma$-labelings $L$ on $\AF$, $L(\cA) = \argin$, so for all $S \in \Ostrat{\sigma}(\AF)$, $S(\cA) = 0 \leq S(\cB)$ for all $\cB \in \arguments$. 
\end{proof}
\end{proposition}

The philosophy pursued in \cite{AmgoudBenNaimSUM2013} is that  \emph{attacks always weaken an argument} whereas in our framework, we aim at assessing the controversiality of an argument, i.\,e., arguments that are clearly defended by other arguments are as uncontroversial. However, if one wishes to do so, a modification of the definition of stratified labelings would be possible where each layer is split into two layers, one (lower) layer containing the non-attacked arguments (which are trivially $\argin$) and a (higher) layer that contains the rest of the $\argin$-Arguments. 

\begin{definition}
Let $\AF= (\arguments,\attacks)$ be an argumentation framework, let $\cA \in \arguments$. Then 
$Att_{\AF}(\cA) = \{\cB \in \arguments \mid \cB \attacks \cA\}$ is the set of \emph{attackers of $\cA$}, and 
$Def_{\AF}(a) = \{\cB \in \arguments \mid \exists \cC \in \arguments \ \mbox{such that} \ \cC \attacks \cA \ \mbox{and} \ \cB \attacks \cC\}$ is the set of \emph{defenders of $\cA$}. 
\end{definition}

\begin{description}
\item{\textbf{Defense Precedence$^*$ \DPstar}} An ordinal semantics $\cO$ satisfies \DPstar\ iff for all argumentation frameworks $\AF= (\arguments,\attacks)$, for all $\lambda \in \cO(\AF)$, for all $\cA, \cB \in \arguments$ the following holds: If $|Att(\cA)| = |Att(\cB)|$ and $Def(\cA) = \emptyset$, but $Def(\cB) \neq \emptyset$, then $\lambda(\cA) < \lambda(\cB)$. 
\end{description}

This postulate is highly debatable as it focusses too much on quite local topological aspects of the argumentation frameworks, in particular, the sheer number of attackers, but neglects the global topology. In our framework, it is more the depth of attacks and the complexity of the topology of the networks that count.  For the same reason, also the postulates \textit{(Strict) Counter-Transitivity}, \textit{Cardinality Precedence}, and \textit{Distributed-Defense Precedence} from \cite{AmgoudBenNaimSUM2013} are not useful in our framework.

The last property from \cite{AmgoudBenNaimSUM2013} to be considered here is \emph{Quality Precedence}: 
\begin{description}
\item{\textbf{Quality Precedence$^*$ \QPstar}} An ordinal semantics $\cO$ satisfies \QPstar\ iff for all argumentation frameworks $\AF= (\arguments,\attacks)$, for all $\lambda \in \cO(\AF)$, for all $\cA, \cB \in \arguments$ the following holds: If there is $\cC \in Att_{\AF}(\cB)$ such that for all $\cD \in Att_{\AF}(\cA), \lambda(\cC) < \lambda(\cD)$, then $\lambda(\cA) < \lambda(\cB)$. 
\end{description}

\textbf{Quality Precedence$^*$} is not satisfied in general by the $\sigma$-ordinal semantics. For instance, as a counterexample, consider Example \ref{Example:GSL3} with $\cA = \cA_5, \cB = \cA_2$. Here we have $Att_{\AF}(\cA) = \{\cA_4\}, Att_{\AF}(\cB) = \{\cA_1\}$, and indeed, $S^{gr}_{\AF}(\cA_1) = 0 < 1 = S^{gr}_{\AF}(\cA_4)$, but $S^{gr}_{\AF}(\cA_5) = 2 > 1 = S^{gr}_{\AF}(\cA_2)$. 

In general, the approach in \cite{AmgoudBenNaimSUM2013} differs from ours in various respects: First, we consider classes of ordinal rankings for argumentation frameworks and not just one (more general) ranking. Second, those authors define a ranking-based semantics in order to assess the strength of an argument while we aim at assessing the controversiality of an argument. Other properties might be more useful and it is up to future work to develop and investigate such properties.

\section{Relating Stratified Labelings with the $(\sigma,\mathcal{U})$-characteristic}\label{sec:rel:char}
In \cite{Baumann:2012} Baumann investigates how arguments can be enforced to be accepted by minimal changes of the underlying argumentation framework. The core notion of his framework is the $(\sigma,\mathcal{U})$-characteristic which is based on the \emph{attack-distance} of two frameworks.

\begin{definition}
	Let $\AF_{1}=(\arguments_{1},\attacks_{1}),\AF_{2}=(\arguments_{2},\attacks_{2})$ be abstract argumentation frameworks. The \emph{attack-distance} $d(\AF_{1},\AF_{2})$ between $\AF_{1}$ and $\AF_{2}$ is defined via 
	\begin{align*}
		d(\AF_{1},\AF_{2})=|\attacks_{1}\Delta \attacks_{2}|\qquad.
	\end{align*}
\end{definition}

\begin{definition}
	Let $\AFcomplete$ be an abstract argumentation framework and $\sigma$ a semantics. The $(\sigma,\mathcal{U})$-characteristic $N^{\AF}_{\sigma}(C)$ of a set $C\subseteq \arguments$ is defined as
	\begin{align*}
		N^{\AF}_{\sigma}(C) & = \left\{\begin{array}{ll}
										0	& C\subseteq \argin(L)\text{~for some~}\sigma\text{-labeling~}L\\
										k 	& k=\min\{d(\AF,\AF')\mid N^{\AF'}_{\sigma}(C) = 0\}\\
										\infty & \text{otherwise}
									\end{array}\right.
	\end{align*}
\end{definition}

The $(\sigma,\mathcal{U})$-characteristic of a set of arguments $C$ is the minimal effort required to establish $C$ being accepted. 

\begin{conjecture}
\label{conj_baumann}
	Let $\AFcomplete$ be an argumentation framework without cycles and $\sigma$ a semantics. For all $\cA,\cB\in\arguments$, if $N^{\AF}_{\sigma}(\{\cA\})<N^{\AF}_{\sigma}(\{\cB\})$ then for all finite $\sigma$-stratified labelings $S$ it holds $S(\cA)<S(\cB)$.
\end{conjecture}

The general statement does not hold as the following example shows.

\begin{example}\label{Example:GSL4}
The grounded-stratified labeling for the argumentation framework depicted in Figure~\ref{fig:GSL4} is $S^{gr}_{\AF}$ with
	\begin{align*}
		S^{gr}_{\AF}(\cA_{1}) &=  2&
		S^{gr}_{\AF}(\cA_{2}) &=  1&
		S^{gr}_{\AF}(\cA_{3}) &=  \infty \\
		S^{gr}_{\AF}(\cA_{4}) &=  \infty &
		S^{gr}_{\AF}(\cA_{5}) &= 1 &
		S^{gr}_{\AF}(\cA_{6}) &= 0 \\
		S^{gr}_{\AF}(\cA_{7}) &= 1 &
		S^{gr}_{\AF}(\cA_{8}) &= 0
	\end{align*}
	but we have $N^{\AF}_{gr}(\{\cA_{1}\})=1$ (by just removing the attack from $\cA_{2}\attacks\cA_{1}$) and $N^{\AF}_{gr}(\{\cA_{2}\})=2$ (by removing e.\,g.\ the attacks $\cA_{4}\attacks \cA_{2}$ and $\cA_{3}\attacks \cA_{2}$).
	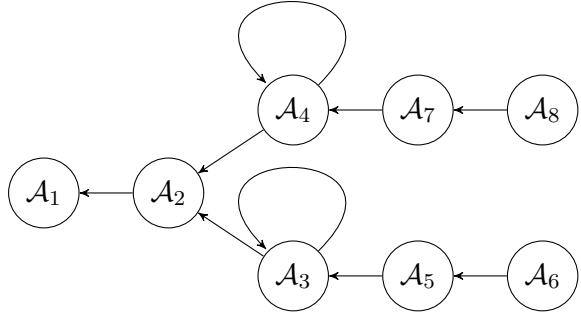
\begin{figure}[h]
		\begin{center}
			\begin{tikzpicture}[node distance=0.7cm]
			
				\node[args](args1){$\cA_{1}$};
				\node[args, right=of args1](args2){$\cA_{2}$};
				\node[args, right=of args2, yshift=-1.1cm](args3){$\cA_{3}$};
				\node[args, right=of args2, yshift=1.1cm](args4){$\cA_{4}$};
				\node[args, right=of args3](args5){$\cA_{5}$};
				\node[args, right=of args5](args6){$\cA_{6}$};
				\node[args, right=of args4](args7){$\cA_{7}$};
				\node[args, right=of args7](args8){$\cA_{8}$};
			
				\path(args8) edge [->] (args7);
				\path(args7) edge [->] (args4);
				\path(args6) edge [->] (args5);
				\path(args5) edge [->] (args3);
				\path(args3) edge [->] (args2);
				\path(args4) edge [->] (args2);
				\path(args2) edge [->] (args1);
				\path(args4) edge [->, loop] (args4);
				\path(args3) edge [->, loop] (args3);

			\end{tikzpicture}
		\end{center}
		\caption{Argumentation framework from Example~\ref{Example:GSL4}}
		\label{fig:GSL4}
	\end{figure}
\end{example}

\section{Further Works}\label{sec:related}
Although \cite{Kern-Isberner:2011} also investigates the relationship between argumentation and ordinal conditional functions, in general, and system Z, in particular, the methods used in that paper are quite different from the approach presented here. In \cite{Kern-Isberner:2011}, the arguments are built from rules, and the argumentation framework suitably chosen there is \textsc{DeLP} \cite{GarciaSimari04}. In the present approach, arguments are more atomic (i.\,e., possible worlds), and the argumentation semantics are abstract. The relations between system Z and argumentation could only be established in some special cases in \cite{Kern-Isberner:2011}, whereas we found a general argumentative characterization of system Z here. 

In \cite{Weydert:2012}, Weydert defines so-called \emph{ranking models} for abstract argumentation frameworks. He associates a kind of conditional with each argument, symbolizing premise and claim of the argument, and interprets attack in terms of (generalized) ordinal conditional functions. While such functions are also the basis for defining system Z, the contributions of that work are quite different from our approach. Most prominently, we assign ranking degrees to abstract arguments, not to the propositional content of arguments. Moreover, in our framework, these ranking degrees are computed solely on the base of the abstract topological structure of the argumentation graph whereas in \cite{Weydert:2012}, rankings are induced partly by the conditionals associated with the arguments, i.\,e., by the internal structures of the arguments. 

\section{Summary and conclusion}\label{sec:summary}
In this paper, we presented preliminary work on a novel semantical notion for abstract argumentation frameworks: stratified labelings. We analyzed some general properties of our approach and compared it to ranking functions for conditional reasoning and similar approaches from argumentation theory. The core difference between our approach of stratified labelings and other \emph{graded} approaches for semantics is that we measure controversiality of arguments instead of their strength.

Ongoing work is about a deeper analysis of the approach and its relationships to other approaches.

\bibliographystyle{alpha}
\bibliography{references}

\end{document}